\documentclass{article}

\usepackage{arxiv}

\usepackage[utf8]{inputenc}
\usepackage[T1]{fontenc}
\usepackage{hyperref}   
\usepackage{url}        
\usepackage{booktabs}   
\usepackage{amsfonts}   
\usepackage{nicefrac}   
\usepackage{microtype}  
\usepackage{cleveref}   
\usepackage{lipsum}     
\usepackage{graphicx}
\usepackage{natbib}
\usepackage{doi}

\title{RCMHA: Relative Convolutional Multi-Head Attention for Natural Language Modelling}

\newif\ifuniqueAffiliation
\uniqueAffiliationtrue

\ifuniqueAffiliation 
\author{ 
    \href{https://orcid.org/0000-0003-4610-0456}{\includegraphics[scale=0.06]{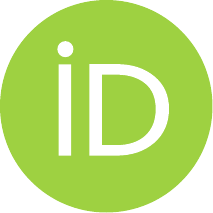}\hspace{1mm} Herman Sugiharto} \\
	Department of Informatics\\
	Siliwangi University\\
	Tasikmalaya, Indonesia \\
	\texttt{177006045@student.unsil.ac.id} \\
	\And
	\href{ https://orcid.org/ 0000-0003-2193-7110}{\includegraphics[scale=0.06]{orcid.pdf}\hspace{1mm} Aradea} \\
	Department of Informatics\\
	Siliwangi University\\
	Tasikmalaya, Indonesia \\
	\texttt{aradea@unsil.ac.id} \\
	\AND
	Husni Mubarok \\
	Department of Informatics\\
	Siliwangi University\\
	Tasikmalaya, Indonesia \\
	\texttt{husni.mubarok@unsil.ac.id} \\
}
\else
\usepackage{authblk}

\newbox{\orcid}\sbox{\orcid}{\includegraphics[scale=0.06]{orcid.pdf}} 
\author[1]{
}
\author[1,2]{%
	\href{https://orcid.org/0000-0000-0000-0000}{\usebox{\orcid}\hspace{1mm}Elias D.~Striatum\thanks{\texttt{stariate@ee.mount-sheikh.edu}}}%
}
\affil[1]{Department of Computer Science, Cranberry-Lemon University, Pittsburgh, PA 15213}
\affil[2]{Department of Electrical Engineering, Mount-Sheikh University, Santa Narimana, Levand}
\fi

\renewcommand{\headeright}{RCMHA}

\hypersetup{
pdftitle={RCMHA: Relative Convolutional Multi-Head Attention for Natural Language Modelling},
pdfsubject={q-cs.AI, q-cs.AI},
pdfauthor={Herman Sugiharto, Aradea, Husni Mubarok},
pdfkeywords={attention module, language modelling, Attention},
}

\begin{document}
\maketitle

\begin{abstract}
	The Attention module finds common usage in language modeling, presenting distinct challenges within the broader scope of Natural 
	Language Processing. Multi-Head Attention (MHA) employs an absolute positional encoding, which imposes limitations on token length 
	and entails substantial memory consumption during the processing of embedded inputs. The current remedy proposed by researchers 
	involves the utilization of relative positional encoding, similar to the approach adopted in Transformer-XL or Relative Multi-Head Attention 
	(RMHA), albeit the employed architecture consumes considerable memory resources. To address these challenges, this study endeavors to refine MHA, 
	leveraging relative positional encoding in conjunction with the Depth-Wise Convolutional Layer architecture, which promises heightened accuracy 
	coupled with minimized memory usage.

	The proposed RCMHA framework entails the modification of two integral components: firstly, the application of the Depth-Wise Convolutional Layer 
	to the input embedding, encompassing Query, Key, and Value parameters; secondly, the incorporation of Relative Positional Encoding into the attention 
	scoring phase, harmoniously integrated with Scaled Dot-Product Attention. Empirical experiments underscore the advantages of RCMHA, wherein it exhibits 
	superior accuracy, boasting a score of 0.572 in comparison to alternative attention modules such as MHA, Multi-DConv-Head Attention (MDHA), and RMHA. 
	Concerning memory utilization, RMHA emerges as the most frugal, demonstrating an average consumption of 2.98 GB, surpassing RMHA which necessitates 3.5 GB.
\end{abstract}


\section{Introduction}

Natural Language Modeling (LM) exhibits distinct characteristics and challenges that set it apart from other sub-fields within the domain of Natural Language Processing. 
LM is concerned with the processing of two fundamental language components: tokens and grammar, as pointed out by Kumar \cite{kumar2019calibration}. 
During the treatment of these components, the LM model is tasked with encoding the sequence of words into a vector, a process crucial for subsequent computations within the neural network, 
thus enabling the model to glean the underlying semantics of each word \cite{Vathsala2020}. Within the realm of LM, a foundational quandary arises: the need to prioritize words that warrant 
immediate processing attention. Bahdanau's research \cite{bahdanau2016neural} presents potential remedies for this issue through the application of an attention mechanism, enabling the LM model 
to ascertain and rank relevant words deserving of focused scrutiny.

The utility of the attention mechanism extends beyond Language Modeling (LM) and finds application in various domains, 
including computer vision for tasks such as filter focusing \cite{wang2018nonlocal} \cite{10.5555/2969033.2969222}, feature channel calibration \cite{Hu2020}, 
text matching \cite{sukhbaatar2015endtoend} \cite{Cui2017} \cite{kim2017structured}, and Neural Machine Translation \cite{liu-etal-2016-neural} \cite{mi-etal-2016-supervised}. 
The amalgamation and customization of attention mechanisms yield the creation of attention modules. One prominent instance is Multi-Head Attention (MHA) \cite{vaswani2023attention}, 
which modularizes the attention mechanism by employing multiple parallel attention computations and subsequently integrating them through Scaled Dot-Product attention.

Progressions within the realm of Multi-Head Attention have been observed across multiple research endeavors, as exemplified by Wang's work \cite{wang2020linformer}, 
wherein MHA has been adapted into Multi-Head Linear Attention or Linformer. This adaptation introduces two linear projection matrices into the key and value calculations. 
In the context of this study, Multi-DConv-Head Attention (MDHA) \cite{so2022primer} extends the framework further by incorporating a 3x1 Depthwise Convolution Layer into the query, key, and value computations subsequent to their projection.

The exploration of Relative Multi-Headed Attention, as outlined by Dai \cite{dai2019transformerxl}, is characterized by the replacement of the conventional absolute positional encoding, 
commonly employed in attention-based assessments, with relative positional decoding. This alteration permits the expansion of input tokens beyond the limitations imposed by absolute positional encoding, 
thereby fostering unlimited input token capacity.

This research endeavors to enhance Multi-Head Attention (MHA) through the incorporation of relative positional encoding. 
Conventionally, MHA relies on absolute positional encoding, which imposes a constraint on the token length that the model can effectively handle. 
This token length limitation curtails the extent of input processing and, consequently, may lead to a reduction in accuracy. 
On the other hand, Relative Multi-Headed Attention (RMHA) excels in accuracy by adopting a relative positional encoding approach; 
however, it grapples with memory usage concerns.

To address the limitations associated with RMHA while preserving its accuracy advantages, this study will adopt a strategic approach. 
Specifically, the Depth-Wise Convolution technique, previously employed in Multi-DConv-Head Attention (MDHA), 
will be integrated into the enhanced MHA framework. The central components of the attention module, namely the query, key, and value inputs, 
will undergo both projection and convolution processes. These processes are designed to achieve a twofold goal: to optimize memory utilization 
and to bolster the model's ability to capture intricate patterns within the data.

By employing the tandem strategies of relative positional encoding for attention scoring and depth-wise convolution applied to attention inputs prior to attention scoring, 
the aspiration is to achieve a dual objective: attaining a commendable accuracy score while simultaneously curbing memory consumption. 
This combined architecture, christened as Relative Convolutional Multi-Head Attention (RCMHA), holds promising potential for integration into the Natural Language Processing domain, 
particularly within the ambit of Language Modeling.

The amalgamation of these two methodologies anticipates a synergy that can bolster the efficiency and effectiveness of the attention module. 
This holistic approach not only showcases the potential for advancing accuracy in language-related tasks but also underscores the significance of optimized memory utilization—an 
imperative factor for enhancing the practicality and scalability of the model within real-world applications.

\section{Related Work}
Several studies on attention modules and attention mechanisms have been carried out previously, which resulted in various attention modules 
such as the Cross Attention Module by Chen et al., \cite{chen2021crossvit}, Free Transformer by Zhai et al., \cite{zhai2021attention}, research Locatello et al.,
with Slot Attention \cite{locatello2020objectcentric}, Feedback Memory proposed by Fan et al., \cite{fan2021addressing}, and Graph Self-Attention by Lavril et al., \cite{ye2019bptransformer}. 
This study will focus on developing Multi-Head Attention as in the research conducted by So et al ., \cite{so2022primer}, 
who projected attentional inputs and used Depth-Wise Convolution on projected inputs. For this reason, this research is known as Multi D-Conv-Head Attention 
because it adds Layer Depth-Wise Convolution. Research has not produced a high accuracy score compared to ordinary Transformers. 
Depth-Wise Convolution focuses on reducing the amount of memory usage as well as research by Merity about Single Headed Attention 
RNN: Stop Thinking With Your Head \cite{merity2019single}, Zhai et al., with An Attention Free Transformer \cite{zhai2021attention}, Child et al., who proposed 
Generating Long Sequences with Sparse Transformers \cite{child2019generating}, and Ye et al., who proposed a study entitled BP-Transformer: Modeling 
Long-Range Context via Binary Partitioning \cite{ye2019bptransformer}.

In this study, the memory reduction process is implemented using Depth-Wise Convolution; this is due to the ease of implementation and appropriate use by 
implementing the projection and convolutional layers on the input, namely Query, Key and Value. Depth-Wise Convolution, which has a deficiency in accuracy, 
will be corrected using other methods focused on the attention scoring section.

Research that can improve the accuracy of the attention module and correct the shortcomings of Depth-Wise Convolution as conducted by Dai et al., \cite{dai2019transformerxl}
in the Transformer-XL study: Attentive Language Models Beyond a Fixed-Length Context. Researchers use relative positional encodings compared to absolute positional 
encodings because the absolute positional encoding used in attention scoring on Multi-Head Attention requires extensive resources. 
After all, it processes all existing segments. Named research that improves accuracy such as Woo et al., with CBAM: Convolutional Block Attention Module \cite{woo2018cbam}, 
Wang et al., through Pyramid Vision Transformer: A Versatile Backbone for Dense Prediction without Convolutions \cite{wang2021pyramid}, and Wang et al ., who proposed 
a method called Linformer: Self-Attention with Linear Complexity \cite{wang2020linformer}. Named studies focus on modifying the attention layer, which has the potential to reduce 
performance due to large memory. Transformer XL, which uses relative positional encodings, is the module used in this research because it directly modifies 
attention scoring without adding layers.

\section{Methodology}
The primary innovation sought by this proposed research lies in the modification of the Multi-Head Attention (MHA) architecture. 
This modification entails the fusion of two key components: the integration of a depth-wise convolutional layer and the incorporation of relative positional encoding. 
The aim is to synthesize these elements to create an advanced architecture that can enhance the performance of the attention mechanism within the context of Natural Language Processing.

\begin{figure}
	\centering
	\includegraphics[width=0.5\textwidth]{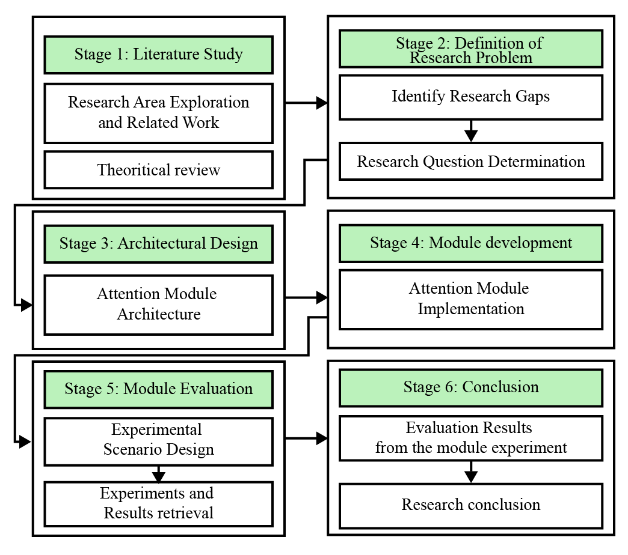}
	\caption{Research stages.}
	\label{fig:fig1}
\end{figure}

The research stages outlining the progression of this study are comprehensively illustrated in Figure \ref{fig:fig1}. This graphical representation encapsulates the sequential steps and 
stages that will be undertaken to realize the proposed architecture, highlighting the logical flow of the research process.

\paragraph{Literature Review}
The preliminary phase involves an extensive examination of existing literature to grasp the fundamental concepts and theories germane to the research domain. 
Key areas of focus encompass theories underpinning Language Modeling, Attention mechanisms, Multi-Head Attention, Relative Positional Encodings, and Depth-Wise Convolution. 
This process entails thorough information retrieval from secondary sources including web resources, academic journals, e-books, articles, and various other scholarly materials.

Furthermore, within the literature review stage, an additional endeavor entails conducting a comprehensive "review paper" or an analysis of prior research endeavors intimately 
aligned with the research subject. This scrutiny of preceding works serves to establish a comprehensive foundation and contextual backdrop, aiding in the identification of gaps, opportunities, and potential directions for the forthcoming research.

\paragraph{Problem Formulation}
Building upon the insights gleaned from the reviewed literature, the subsequent step involves delineating the research problem. 
This entails pinpointing the gaps or limitations inherent in the prior studies, thereby facilitating the identification of areas that warrant further investigation or enhancement. 
By meticulously identifying these gaps or inadequacies, a foundational basis is established for driving improvements. Following this identification process, research questions are formulated, 
aligning with the intention of achieving the defined research objectives.

\paragraph{Architectural Conceptualization}
Progressing to the architectural design phase, meticulous attention is devoted to crafting the Relative Convolutional Multi-Head Attention architecture. 
This architectural blueprint is designed to encapsulate the integration of relative positional encoding and depth-wise convolutional techniques. 
The design is rendered in the form of both mathematical models and visual diagrams, affording a comprehensive representation of the proposed architecture's fundamental components and interactions. 
This design phase serves as a precursor to the subsequent implementation and empirical evaluation stages of the research.

\paragraph{Module Evaluation}
Proceeding to the module evaluation phase, the meticulously crafted attention module will undergo rigorous testing via a series of meticulously designed experiments. 
This experimental regimen encompasses the execution of the model across a spectrum of diverse parameters, including $d_{model}$, number of heads, and $p_{drop}$ (dropout probability). 
Subsequently, a meticulous comparison is undertaken between the attention module with optimal parameters and other prevailing attention modules. 
This comparative analysis seeks to discern the module's performance advantages over its counterparts.

The resultant outcomes stemming from the conducted experiments are methodically elucidated and presented within this section. 
The presentation encompasses a detailed exposition of the experimental findings, supplemented by the articulation of key insights and trends. 
The presentation format incorporates both tabular representations and graphical diagrams, which collaboratively convey the quantitative outcomes of the experiments, 
ensuring a comprehensive and lucid depiction of the module's performance.

\paragraph{Conclusion}
The research culminates in a definitive conclusion, encapsulating a comprehensive overview of the conducted data analysis and the resultant module evaluation outcomes. 
This concluding segment provides a coherent synthesis of the research journey, encapsulating critical aspects of the undertaken investigation. 
It distills the findings and insights garnered from the research, offering a concise perspective on the efficacy and potential implications of the proposed Relative 
Convolutional Multi-Head Attention architecture within the realm of Natural Language Processing.

\section{Results and Discussion}

\subsection{Architectural Developments}
\begin{figure}
	\centering
	\includegraphics[width=0.5\textwidth]{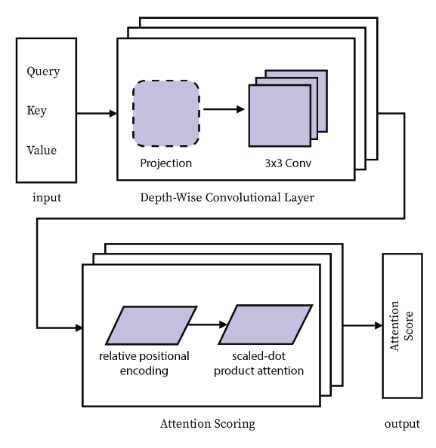}
	\caption{Overview of RCMHA architecture.}
	\label{fig:fig2}
\end{figure}

The architecture of the Relative Convolutional Multi-Head Attention (RCMHA) is an amalgamation of two core concepts: the Depth-Wise Convolutional Layer and Relative Positional Encoding (RPE). 
This synthesis is vividly depicted in Figure \ref{fig:fig2}, illustrating the interplay and integration of these pivotal components within the RCMHA framework.

\paragraph{Relative Positional Encoding} 
In conventional Transformers, the calculation of attention scores involving a query vector $q_i$ and a key vector $k_j$ from the same segment is represented as per Equation (1):

\begin{equation}
	A_(i,j)^abs= (E_(x_i)^T W_q^T W_k E_(x_j ) )+ (E_(x_i)^T W_q^T W_k U_j ) + (U_i^T W_q^T W_k E_(x_j ) )+ (U_i^T W_q^T W_k E_j ) 
\end{equation}

In the context of relative positional encoding, this convention undergoes alteration through the substitution of the conventional absolute positional encoding $U_j$ with the novel 
concept of relative positional encoding $R_{i-j}$. The implementation of relative positional encoding entails specific adaptations within the Scaled Dot-Product Attention, 
where the calculation of attention values takes place in the Multi-Head Attention (MHA) context.

\begin{figure}
	\centering
	\includegraphics[width=0.5\textwidth]{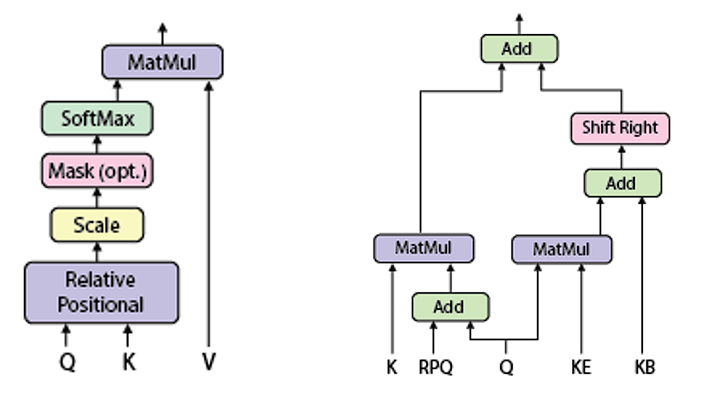}
	\caption{Left: Attention Scoring \cite{vaswani2023attention}, Right : Relative Positional Encoding \cite{wang2021pyramid}.}
	\label{fig:fig3}
\end{figure}

Figure \ref{fig:fig3} visually elucidates the precise modifications undertaken within the attention score computation mechanism and the integration of relative positional encoding. 
These modifications collectively constitute an integral part of the broader enhancements introduced within the framework.

\paragraph{Depth-Wise Convolutional Layer}
The second pivotal enhancement introduced to the Multi-Head Attention (MHA) involves the integration of a Depth-Wise Convolutional (DWC) layer into each of the constituent input values—namely Query, Key, and Value. 
This augmentation aims to bolster the module's overall capability by infusing an additional layer within the processing pipeline. 
Notably, this modification is accompanied by a transformation in the activation function employed. Specifically, the conventional Rectified Linear Unit (ReLU) activation is supplanted by the employment of the Squared ReLU activation function.

\begin{figure}
	\centering
	\includegraphics[width=0.5\textwidth]{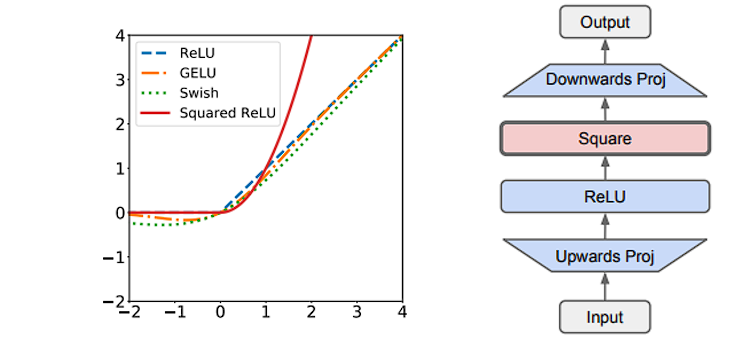}
	\caption{Left: Squared ReLU has different asymptotics with other activation functions. Right: Squared Relu architecture \cite{so2022primer}.}
	\label{fig:fig4}
\end{figure}

Squared ReLU distinguishes itself from other activation functions through its distinctive asymptotic behavior, as illustrated in Figure \ref{fig:fig4}. 
Empirical investigations detailed in this study \cite{so2022primer} underscore the efficacy of Squared ReLU, which is observed to outperform conventional ReLU and its variants— including the standard variant of ReLU. 
These experiments substantiate the superiority of Squared ReLU in terms of its impact on enhancing the performance of the attention module.

\begin{figure}
	\centering
	\includegraphics[height=0.2\textheight]{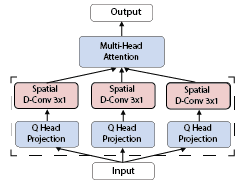}
	\caption{Depth-Wise Convolution \cite{so2022primer}.}
	\label{fig:fig5}
\end{figure}

Undoubtedly, a pivotal modification within this section entails the inception of the Depth-Wise Convolutional (DWC) layer. The DWC layer is crafted through a two-step process: 
initially projecting each individual input value, encompassing Query (Q), Key (K), and Value (V). 
Figure \ref{fig:fig5} intricately illustrates the integration of DWC in conjunction with the Multi-Head Attention (MHA) architecture.

This amalgamation not only amplifies the architectural complexity but also ushers in a transformative enhancement to the overall performance of the module. 
The DWC layer's integration adds a crucial dimension to the module's computational process, thereby contributing to the evolution of its processing capabilities. 
This pivotal augmentation and its graphical representation further underscore the research's innovative approach to refining attention mechanisms within the realm of Natural Language Processing.

\begin{figure}
	\centering
	\includegraphics[width=0.5\textwidth]{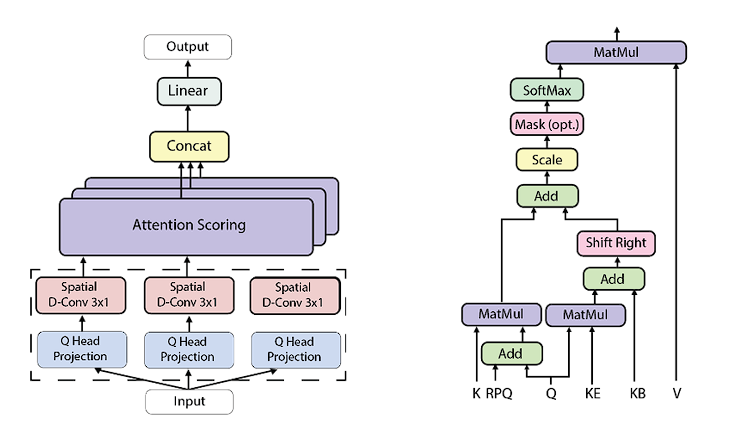}
	\caption{\emph{Relative Convolutional Multi-Head Attention.}}
	\label{fig:fig6}
\end{figure}

\paragraph{Relative Convolutional Multi-Head Attention}
The culmination of these advancements gives rise to the creation of the Relative Convolutional Multi-Head Attention Module (RCMHA), as depicted in Figure \ref{fig:fig6}. 
This innovative module is a result of the synergistic integration of the previously delineated components: Relative Positional Encoding and Depth-Wise Convolutional Layer.

In terms of its functional characteristics, RCMHA aligns with the input-output paradigm of the conventional Multi-Head Attention (MHA). 
The inputs encompass Query (Q), Key (K), and Value (V), while the outputs encompass attention and attention scores. 
Notably, the attention scoring mechanism within RCMHA stands as a fusion of relative positional encoding and scaled-dot production. 
This hybrid mechanism serves as a testament to the intricate and sophisticated nature of RCMHA, aptly capturing the research's innovative approach to amplifying attention mechanisms within the broader landscape of Natural Language Processing.

In module development, several libraries are used to assist its development. Table \ref{tab:table1} is a list of libraries used in development.

\begin{table}
	\caption{Libraries used}
	\centering
	\begin{tabular}{lll}
		\toprule
		\cmidrule(r){1-2}
		No     & Library     & Usage \\
		\midrule
		1	& Pytorch  & Attention module creation     \\
		2	& LabML.ai & Experiment creation      \\
		\bottomrule
	\end{tabular}
	\label{tab:table1}
\end{table}

\begin{figure}
	\centering
	\includegraphics[height=0.2\textheight]{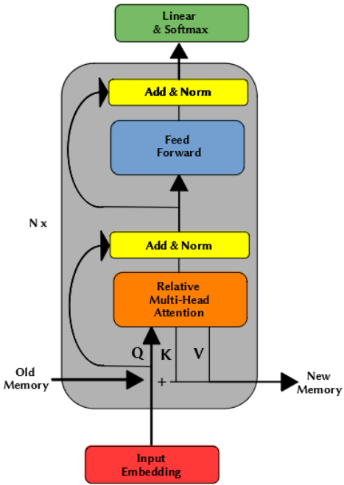}
	\caption{\emph{XL Transformer Architecture \cite{dai2019transformerxl}.}}
	\label{fig:fig7}
\end{figure}

In practical application, the utilization of the Attention Module necessitates the presence of a model framework capable of accommodating data training; 
this framework is commonly referred to as a transformer. In the context of this research, the chosen transformer model is TransformerXL. 
This specific model is particularly tailored to accommodate the intricacies of Relative Multi-Head Attention, as aptly illustrated in Figure \ref{fig:fig7}.

The selection of TransformerXL underscores the deliberate alignment of the model with the research's focus on Relative Multi-Head Attention. 
By leveraging this model, the research harnesses an architecture designed to synergistically integrate the advancements detailed within the research, 
thus paving the way for robust experimentation and evaluation within the context of Natural Language Processing tasks.

\subsection{Module Experiment Design}
This study is poised to execute a comprehensive set of experiments aimed at determining optimal parameters and conducting thorough comparisons with other attention models. 
The core objective of these experiments is to facilitate a quantitative assessment of both memory utilization and accuracy. 
Through these evaluations, the study endeavors to substantiate the viability of specific variants and models that align with the predefined objectives.

This overarching goal is pursued via a series of meticulously designed experiments, each strategically employing a range of variables germane to achieving the stated objectives. 
The experimental parameters, integral to this endeavor, are meticulously outlined in Table \ref{tab:table2}. 
These parameters serve as the foundational basis for systematically exploring the relationships between different configurations and their resultant impact on memory consumption and accuracy levels.

\begin{table}
	\caption{Experimental parameters}
	\centering
	\begin{tabular}{lll}
		\toprule
		\cmidrule(r){1-2}
		Parameter     & Value tested	& Value observed \\
		\midrule
		$d_{model}$  & 128,256,512		& Memory load and accuracy     \\
		$p_{drop}$  & 0, 0.1		& Memory load     \\
		\emph{heads}  & 4, 8		& Memory load and accuracy     \\
		\bottomrule
	\end{tabular}
	\label{tab:table2}
\end{table}

The experimental framework of this study entails the incorporation of each attention module within a consistent training model architecture within the domain of language modeling. 
To establish a benchmark, the AutoRegression model has been selected—a widely employed reference in pertinent research endeavors \cite{dai2019transformerxl}. 
The pivotal component driving the AutoRegression model is the Transformer XL, chosen for its regression capabilities. The attention module, constituting the focus of this study, is seamlessly integrated as the initial element in the input processing pipeline. 
Notably, this module undertakes the embedding of Query, Key, and Value input values.

\begin{figure}
	\centering
	\includegraphics[width=0.5\textwidth]{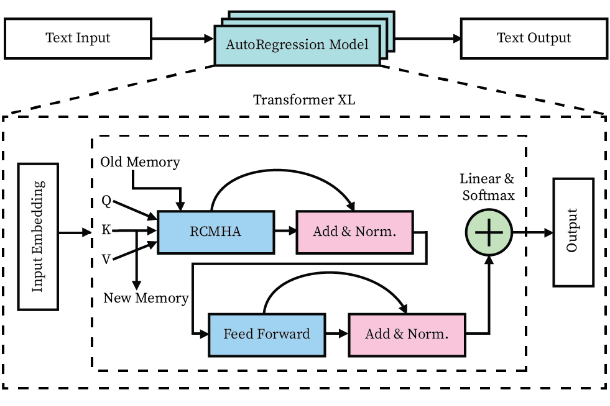}
	\caption{Model training architecture}
	\label{fig:fig8}
\end{figure}

Figure \ref{fig:fig8} vividly illustrates this test architecture, highlighting the strategic positioning of the attention module within the overall framework. 
This architecture not only provides the foundation for conducting meaningful experiments but also underscores the meticulous alignment of research objectives with the chosen methodologies and evaluation strategies.

For the purpose of conducting the experiments pertaining to the attention modules, the LabML.ai library will be harnessed. 
This library serves as a valuable tool for designing and executing modules in a manner that expedites experimentation processes. 
Moreover, Neptune.ai will be leveraged to measure and visualize the performance of the models. By transmitting critical parameters, 
hardware metrics (such as memory and CPU utilization), and key outcomes (accuracy, PPI, and loss) to the Neptune.ai server, 
a comprehensive record is maintained and visualizations are generated for subsequent analysis.

\begin{figure}
	\centering
	\includegraphics[width=0.5\textwidth]{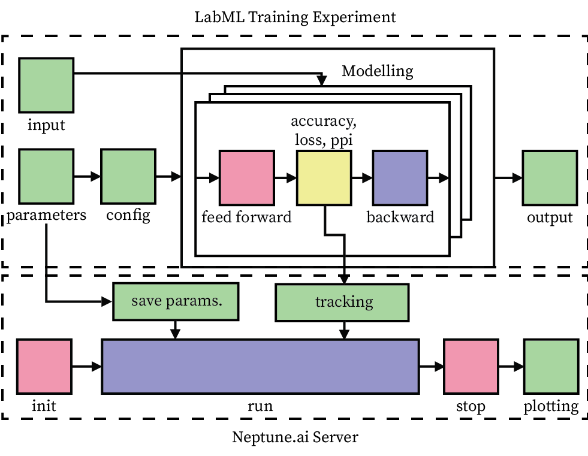}
	\caption{Training and measurement scheme}
	\label{fig:fig9}
\end{figure}

The intricacies of the training scheme, coupled with the measurement procedures, are cogently illustrated in Figure \ref{fig:fig9}. 
This visualization encapsulates the orchestration of the experimental process—ranging from the execution of training cycles to 
the meticulous tracking of performance metrics. The combined utilization of LabML.ai and Neptune.ai underscores the research's commitment 
to methodical experimentation and precise measurement within the realm of Natural Language Processing.

The dataset used in the experiment is tiny shakespeare, an available dataset for testing language modelling. 
Experiments using the help of google collaborator pro with hardware specifications can be seen in Table \ref{tab:table3}.

\begin{table}
	\caption{Hardware specifications}
	\centering
	\begin{tabular}{ll}
		\toprule
		\cmidrule(r){1-2}
		Hardware     & Spec \\
		\midrule
		CPU 	&	2 x Intel Xeon CPU @ 2.20GHz \\
		GPU		&	Tesla P100 16GB \\
		RAM		&	27GB \\
		Storage	&	129GB available \\
		\bottomrule
	\end{tabular}
	\label{tab:table3}
\end{table}

\subsection{Module Evaluation}
During the evaluation phase, a comprehensive comparative analysis will be conducted to juxtapose the performance of various attention modules. 
Specifically, the attention modules under scrutiny include Multi-Head Attention, Multi-DConv-Head Attention, Relative Multi-Head Attention, and the novel contribution 
of this research—Relative Convolutional Multi-Head Attention. 

The evaluation encompasses not only a comparison among distinct attention modules but also an exploration of various module variations achieved 
by altering critical model parameters. These parameters encompass dimensions, the number of heads, and dropout rates. 
By systematically varying these parameters, a comprehensive understanding of the modules' behaviors and their respective impacts on performance can be discerned. 
This meticulous evaluation approach underscores the research's commitment to rigorously scrutinizing the proposed advancements and their implications within 
the realm of Natural Language Processing.

\paragraph{Variation}
The variation of the module used in this study consisted of 8 variations, as shown in Table \ref{tab:table4} below:

\begin{table}
    \caption{RCMHA variations}
    \centering
    \begin{tabular}{llll}
        \toprule
        \cmidrule(r){1-2}
        code & $d_{model}$ & \emph{heads} & $p_{drop}$ \\
        \midrule
        A & 128 & 4 & 0 \\
        B & 128 & 8 & 0 \\
        C & 256 & 4 & 0 \\
        D & 256 & 8 & 0 \\
        E & 256 & 8 & 0.1 \\
        F & 512 & 4 & 0 \\
        G & 512 & 8 & 0 \\
        H & 512 & 8 & 0.1 \\
        \bottomrule
    \end{tabular}
    \label{tab:table4}
\end{table}

\begin{figure}
	\centering
	\includegraphics[width=0.5\textwidth]{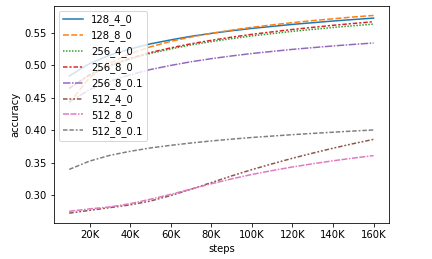}
	\caption{Visualization of module comparison results}
	\label{fig:fig10}
\end{figure}

\begin{figure}
	\centering
	\includegraphics[width=0.5\textwidth]{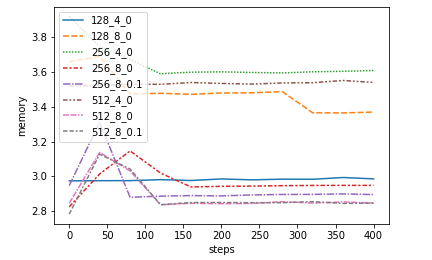}
	\caption{Visualization of variation comparison results}
	\label{fig:fig11}
\end{figure}

From these variations, it can be found that the optimal value for \emph{variation A} with high accuracy but has PPL and low memory usage. 
Tables \ref{tab:table5} illustrate the comparison between the variations and are visualized in Figures \ref{fig:fig10} and \ref{fig:fig11}.

\begin{table}
    \caption{Variation performance result}
    \centering
    \begin{tabular}{llllllll}
        \toprule
        \cmidrule(r){1-2}
        \# & train steps & PPL & acc loss & params & mem (avg) (GB) & CPU (avg) \\
        \midrule
        A & 174269 & 3.77187 & 0.57252 & 1.32757 & 7.30E+06 & 2.98573 & 28.3701 \\
        B & 174269 & 7.36633 & 0.576625 & 199692 & 7.50E+06 & 3.46275 & 28.5655 \\
        C & 174269 & 4.177 & 0.563293 & 1.42959 & 1.52E+07 & 3.59483 & 28.9132 \\
        D & 174269 & 4.72575 & 0.566993 & 1.55303 & 1.54E+07 & 2.94664 & 29.5094 \\
        E & 174269 & 4.60318 & 0.534253 & 1.52675 & 1.54E+07 & 2.92002 & 29.1196 \\
        F & 174269 & 5.67633 & 0.386044 & 1.73631 & 3.33E+07 & 3.53417 & 28.4058 \\
        G & 174269 & 6.1152 & 0.361048 & 1.81078 & 3.35E+07 & 2.86167 & 29.305 \\
        H & 174269 & 6.73797 & 0.40504 & 1.90776 & 3.35E+07 & 2.86255 & 29.2889 \\
        \bottomrule
    \end{tabular}
    \label{tab:table5}
\end{table}

\paragraph{Module Comparison}
The modules used for comparison are Multi-Head Attention which is the basic module of RCMHA. 
Relative Multi-Head Attention is the basis for relative positional encoding, and Multi-DConv-Head Attention is used for Dept-wise Convolutional in RCMHA. 
In this comparison, RCMHA uses its best variation with parameters $d_{model}=128$, $p_{drop}=0$, and $head=4$.

\begin{table}
    \caption{Experimental results of the RCMHA module with the comparison module}
    \centering
    \begin{tabular}{lllll}
        \toprule
        \cmidrule(r){1-2}
        Module & Accuracy & Loss & PPL & Mem. (AVG) \\
        \midrule
        RCM & 0.57252 & 1.32757 & 3.77187 & 2.98573 \\ 
        MHA & 0.566343 & 1.30334 & 3.68157 & 2.93567 \\
        INDIA & 0.557197 & 1.54435 & 4.68493 & 2.92108 \\
        RMHA & 0.555018 & 1.11591 & 3.05233 & 3.50318 \\
        \bottomrule
    \end{tabular}
    \label{tab:table6}
\end{table}

\begin{figure}
	\centering
	\includegraphics[width=0.5\textwidth]{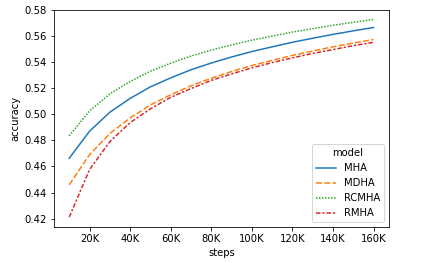}
	\caption{Visualization of module comparison results}
	\label{fig:fig12}
\end{figure}

\begin{figure}
	\centering
	\includegraphics[width=0.5\textwidth]{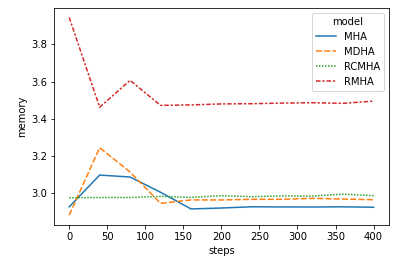}
	\caption{Visualization of module comparison results}
	\label{fig:fig13}
\end{figure}

The comparative analysis presented in Table \ref{tab:table6} and the visualizations provided in Figures \ref{fig:fig12} and \ref{fig:fig13} offer insightful observations. 
Specifically, the Relative Convolutional Multi-Head Attention (RCMHA) module emerges as a standout performer, boasting the highest accuracy score of 0.572. 
This accomplishment is complemented by the added benefit of reduced memory utilization when compared to Relative Multi-Head Attention (RMHA). 

These empirical findings collectively underscore RCMHA's proficiency in addressing RMHA's memory-related challenges, resulting in superior memory efficiency 
while standing on par with the other two modules in terms of memory consumption. Furthermore, RCMHA's architecture, fortified with relative positional encoding, 
exhibits the capability to surmount the accuracy and loss issues encountered by Multi-DConv-Head Attention (MDHA). 
This transformative impact is reflected in the superior accuracy, loss, and Perplexity (PPL) values demonstrated by RCMHA. 
Overall, the study underscores the efficacy of RCMHA in overcoming existing limitations and enhancing performance across various dimensions.

\subsection{Threats of validity}

\paragraph{Internal Threats}
The conducted experiment focused exclusively on the Tiny Shakespeare dataset, which serves as a standard benchmark for language modeling evaluations. 
While this dataset provides valuable insights into the attention module's performance within the context of smaller-scale tasks, 
it might not comprehensively reflect the module's capabilities in handling extensive datasets or high data volumes. 

It is acknowledged that the limitations in equipment and resources influenced the choice of dataset. Future research endeavors can be directed towards 
leveraging more robust computational capabilities to assess the attention module's response to stress levels posed by larger datasets. 
Additionally, the evaluation of the Transformer-XL model, integral to this study, on substantial volumes of data can provide a more comprehensive understanding of its scalability and performance. 
The progression to more extensive and diverse datasets will undoubtedly contribute to a holistic assessment of the proposed advancements and their broader applicability.

\paragraph{External Threats}
Indeed, the execution of experiments within the context of a virtual machine, such as the one employed in Google Colab, can introduce variations in measurements and outcomes. 
The measurements obtained in one virtual machine environment may not directly translate to another machine due to factors such as hardware specifications, memory usage, and the virtualization setup. 

Google Colab's virtual machines are subject to resource sharing, whereby a single hardware device is allocated to multiple virtual machines. 
This sharing mechanism can potentially influence the performance and measurements obtained during experiments. 
Moreover, the inherent limitations of virtualization can impact the precision of performance evaluations.

In light of these considerations, future research initiatives are encouraged to explore the use of dedicated virtual machines with enhanced specifications. 
Such an approach can offer a more controlled environment, allowing for more accurate and consistent performance measurements. 
By mitigating the potential confounding factors introduced by shared resources, researchers can ensure more reliable and generalizable results across different machines and environments.

\section{Conclusions}
The incorporation of relative positional encoding within the framework of Relative Multi-Head Attention (RMHA) indeed yields a notable increase in accuracy. 
However, this accuracy enhancement is accompanied by a trade-off in terms of memory utilization, as evidenced by the experimental results. 
This memory overhead can be mitigated through the strategic inclusion of a depth-wise convolutional layer. 
The outcomes of the experiments incontrovertibly validate the superior performance of the Relative Convolutional Multi-Head Attention (RCMHA) module, which achieves an impressive accuracy score of 0.57252.

Importantly, RCMHA doesn't merely excel in accuracy, but it also demonstrates commendable memory efficiency, with an average usage of 2.98 gigabytes—significantly lower than the memory consumption of RMHA. 
The interplay between accuracy and memory efficiency bears a direct relationship with training speed and time. This relationship is well-illustrated by the training times: RCMHA, 
despite its higher accuracy, requires 2 hours and 2 minutes for training, while RMHA and MDHA achieve shorter training times of 1 hour and 27 minutes and 1 hour and 19 minutes, respectively.

These findings collectively underscore RCMHA's prowess in achieving a balanced optimization between accuracy, memory utilization, and training efficiency—ultimately positioning it as a promising 
innovation within the landscape of attention mechanisms in Natural Language Processing.

\section{Future Work}
This research requires a more comprehensive implementation and experimental proof, for example, by implementing Neural Machine 
Translation or Text Generation. The dataset can also be a projection for further research by using more varied datasets and testing 
in isolated systems to make the performance measurement more accurate. Architectural modifications can also be made to increase the 
performance or speed of the training time, which is the lack of the RCMHA attention module.

\bibliographystyle{unsrtnat}
\bibliography{references}

\end{document}